\documentclass{article}
\usepackage{iclr2027_conference,times}
\usepackage{amsmath,amsfonts}
\usepackage[hyperfootnotes=false]{hyperref}
\usepackage{url}
\usepackage{graphicx}
\usepackage{amssymb}
\usepackage{booktabs}
\usepackage{subcaption}
\usepackage[table]{xcolor}
\usepackage[normalem]{ulem}
\usepackage{hyperref}
\title{Agentic Physics-Grounded Video Generation via Memory-Augmented In-Context Reasoning}
\author{
\textbf{Xikang Guan\textsuperscript{1,\(\dagger\)}},\quad
\textbf{Shaocong Yang\textsuperscript{1,\(\dagger\)}},\quad
\textbf{Yanqi Bao\textsuperscript{2,\(\dagger,\ddagger\)}},\quad
\textbf{Can Wang\textsuperscript{3}} \\
\textbf{Zheng Gu\textsuperscript{4}},\quad
\textbf{Zhengqing Chen\textsuperscript{2}},\quad
\textbf{Kai Chen\textsuperscript{1}},\quad
\textbf{Jing Huo\textsuperscript{1,*}},\quad
\textbf{Yang Gao\textsuperscript{1}} \\[4pt]
{\small\normalfont \textsuperscript{1}State Key Laboratory for Novel Software Technology, Nanjing University} \\
{\small\normalfont \url{{xikangguan,shaocongyang,kaichennju}@smail.nju.edu.cn, {huojing,gaoy}@nju.edu.cn}} \\
{\small\normalfont \textsuperscript{2}Horizon Robotics} \\
{\small\normalfont \url{yanqibao1997@gmail.com, zhengqing.chen@horizon.cc}} \\
{\small\normalfont \textsuperscript{3}School of Computing and Data Science, The University of Hong Kong} \\
{\small\normalfont \url{canwang@hku.hk}} \\
{\small\normalfont \textsuperscript{4}College of Computer Science and Software Engineering, Shenzhen University} \\
{\small\normalfont \url{guzheng@szu.edu.cn}} \\
}

\iclrfinalcopy
\renewcommand{\headrulewidth}{0pt}

\begin{document}
\maketitle
\fancyhead{}

\begingroup
\renewcommand{\thefootnote}{\ensuremath{\dagger}}
\footnotetext{Equal contribution}
\renewcommand{\thefootnote}{\ensuremath{\ddagger}}
\footnotetext{Project leader}
\renewcommand{\thefootnote}{*}
\footnotetext{Corresponding author}
\endgroup

\begin{abstract}
Physical plausibility is essential for realistic video generation.
Existing methods impose physical guidance before synthesis via simulations or refined prompts, but this can limit dynamic diversity, miss errors that emerge only after generation, and require repeated regeneration when inconsistencies remain.
We therefore study a \textit{generate--diagnose--repair} paradigm that preserves the original generation capability and selectively repairs physical inconsistencies afterward, raising two key questions: what to repair and how to repair? To address them, we introduce \textbf{PhysRepair}, an agentic video generation framework that combines causal reasoning over the observed physical failure with prior repair experience to enable localized and adaptive correction. Specifically, \textbf{Spatiotemporal Causal Constraints} identify the first invalid physical state and its causally affected states, producing a repair specification that guides physically plausible local regeneration while preserving unaffected content. \textbf{Experience-augmented In-context Learning} combines the ongoing repair context with relevant experience from previous repairs, allowing the agent to adapt its repair decisions across diverse physical failures and accumulate completed trajectories for future cases. Experiments on Physics-IQ Verified across solid mechanics, fluid dynamics, optics, thermodynamics, and magnetism demonstrate that PhysRepair consistently improves the physical plausibility of generated videos over existing methods. Our project page is available at \href{https://xxkkxxx.github.io/PhysRepair-Website/}{https://xxkkxxx.github.io/PhysRepair-Website/}.

\end{abstract}

\section{Introduction}

\begin{figure}[t]
    \centering
    \includegraphics[width=\columnwidth]{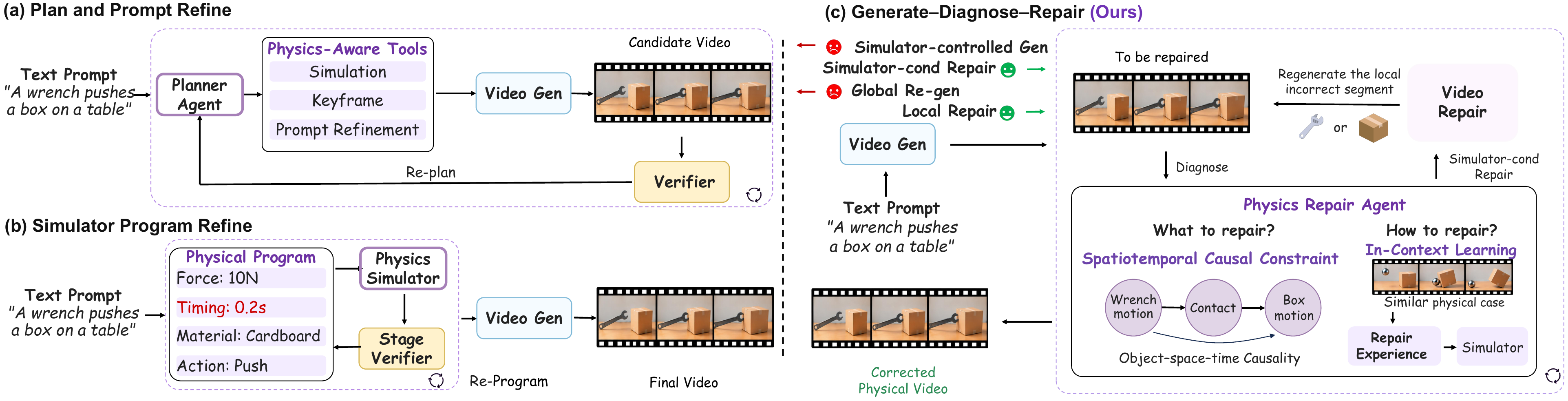}
    \caption{
    \textbf{Comparison of physics-grounded video generation paradigms.}
    \textbf{(a) Plan and Prompt Refine} iteratively updates generation plans or physics-aware prompts, while
    \textbf{(b) Simulator Program Refine} iteratively updates physical programs or simulator configurations; both impose physical guidance before synthesis and may require repeated regeneration.
    \textbf{(c) PhysRepair} instead follows a \textit{generate--diagnose--repair} paradigm: it diagnoses physical failures from the generated video (what to repair?) and performs localized, causally constrained repair, while leveraging prior repair experience to adapt the intervention and preserve generated already-valid content (how to repair?).
    }
    \label{fig:overview}
\end{figure}

Modern video generators achieve impressive visual quality but still struggle with physical plausibility, often producing implausible motion, incorrect contact or collision behavior, and unrealistic responses to physical forces \citep{bansal2024videophy,bansal2025videophy2,guo2025t2vphysbench}.
Existing approaches to physically plausible video generation primarily seek to prevent physical failures during synthesis by incorporating physical knowledge through training data, motion priors, simulation, or iteratively refined physical prompts \citep{liu2024physgen,xue2025phyt2v,yang2025vlipp,nguyen2026physplan}. However, imposing extensive physical guidance before synthesis can constrain the model's ability to generate diverse and complex dynamics, favoring simpler and predefined physical interactions. In contrast, we follow a \textit{generate--diagnose--repair} paradigm that preserves the generator's original synthesis capability and introduces physical guidance only after specific inconsistencies are observed, enabling localized correction of the affected interactions while leaving already-valid content unchanged, as shown in Fig.~\ref{fig:overview}. This formulation raises two coupled challenges: 
1) \textbf{What to repair.} A spatial mask can localize an object, but cannot determine which temporal states should change, whether another object's trajectory should be preserved, or how a physical failure propagates across objects, space, and time \citep{yi2020clevrer,motamed2026void}; 2) \textbf{How to repair.} Diverse physical failures require different physical evidence, simulation configurations, and generation conditions that cannot be predefined for every case, while overly broad corrections may disrupt already-valid interactions and dynamics and introduce new inconsistencies \citep{shinn2023reflexion,xu2026veriphy}.

To address these challenges, we cast physical video repair as an \textbf{adaptive agentic process}: the agent reasons about the causal influence of an observed physical failure to localize the repair to the affected content, while drawing on prior repair processes of related physical failures to determine how the repair should be performed. This enables targeted and adaptive correction without globally modifying already-valid dynamics or solving each failure from scratch. By iteratively diagnosing repaired results and refining subsequent interventions, the agent progressively improves the physical plausibility of the generated video.

Concretely, we introduce \textbf{PhysRepair}, an agentic video generation system that operationalizes this adaptive repair process through \textbf{Spatiotemporal Causal Constraints} and \textbf{Experience-augmented In-context Learning}. For Spatiotemporal Causal Constraints, we organize the diagnosed interaction into an object-centric causal representation, locate the first physically invalid state, and trace its downstream effects to construct a causal repair specification. The specification is then used to reconstruct a physically plausible alternative to the incorrect dynamics and guide localized video regeneration while suppressing conflicting source dynamics. Experience-augmented In-context Learning maintains the evolving repair process in short-term working memory and augments it with relevant repair experience retrieved from long-term experience memory. Together, they provide in-context guidance to adapt agentic repair decisions, while completed trajectories are distilled back into the long-term experience memory for future repairs.

We evaluate PhysRepair on Physics-IQ Verified across five physics categories--solid mechanics, fluid dynamics, optics, thermodynamics, and magnetism--to assess physical plausibility. 
Compared with existing methods, PhysRepair demonstrates improved physical plausibility on Physics-IQ Verified.
Ablation studies further examine the effects of Spatiotemporal Causal Constraints and Experience-augmented In-context Learning.
Our main contributions are:

\begin{itemize}

\item
We introduce \textbf{PhysRepair}, an agentic \textit{generate–diagnose–refine} formulation that iteratively identifies and corrects physical failures after generation, enabling targeted intervention while preserving already-valid dynamics and visual content.

\item
We address the complementary challenges of what to repair and how to repair through two components: Spatiotemporal Causal Constraints and Experience-augmented In-context Learning. The former identifies causally affected object–space–time states to localize intervention while preserving unaffected content, whereas the latter retrieves relevant prior repair experience to adapt repair decisions across diverse physical failures.

\item
We conduct extensive experiments on Physics-IQ Verified across five physics categories--solid mechanics, fluid dynamics, optics, thermodynamics, and magnetism, demonstrating consistent improvements in physical correctness. 

\end{itemize}

\section{Related Work}
\label{sec:related_work}

\subsection{Physics-aware video generation and physical evaluation}

Recent benchmarks reveal that video generators often fail to satisfy physical laws, even when they achieve high photorealism and temporal smoothness. VideoPhy and PhyGenBench evaluate physical common sense across solid, fluid, and multi-domain phenomena \citep{bansal2024videophy,meng2024phygenbench}. Physics-IQ tests whether models can predict physically valid futures from observed experimental frames \citep{motamed2025physicsiq}. Meanwhile, VideoPhy-2 and T2VPhysBench evaluate fine-grained rule adherence and first-principles conservation laws \citep{bansal2025videophy2,guo2025t2vphysbench}. Together, these studies highlight the need to explicitly assess contacts, trajectories, and transitions, which directly provides candidate failure categories for our verifier. Beyond evaluation, another line of work injects physical guidance prior to video synthesis. For example, PhysGen renders rigid-body simulations via diffusion priors \citep{liu2024physgen}. PhyT2V iteratively refines text prompts using physical rules \citep{xue2025phyt2v}. Similarly, VLIPP plans coarse trajectories \citep{yang2025vlipp}, and PhysPlan optimizes motion updates away from passive regions via agentic planning \citep{nguyen2026physplan}. However, imposing extensive physical constraints before synthesis restricts the diversity of generated dynamics, cannot address physical errors that only appear post-generation, and often requires costly global regeneration. In contrast, our \textit{generate--diagnose--repair} paradigm preserves the base model's generative capacity. We introduce physical guidance only after detecting specific inconsistencies, enabling localized corrections while keeping already-valid content intact.

\subsection{Agentic verification, refinement, and experience}

Large language models (LLMs) and vision-language models (VLMs) are increasingly used for physical planning and output evaluation. Beyond pre-generation planning \citep{xue2025phyt2v,nguyen2026physplan}, VeriPhy translates prompts into explicit physical rules, collects tool-based evidence, and verifies outcomes before rewriting prompts \citep{xu2026veriphy}. These systems demonstrate the value of verification feedback, but primarily use it to revise prompts for global regeneration rather than guide localized repair. Moving from global regeneration to localized repair requires not only identifying physical failures but also selecting intervention strategies suited to each case. To support such adaptive decision-making, frameworks such as Reflexion and Voyager offer a useful precedent: they reuse feedback and accumulated experience to guide subsequent actions without updating model weights \citep{shinn2023reflexion,wang2023voyager}, providing a basis for experience-driven in-context learning (ICL). Building on this principle, our Experience-Augmented In-Context Learning combines current repair feedback with relevant prior repair experience in the agent’s context, enabling adaptive repair decisions without solving each failure from scratch.

\section{Method}
\label{sec:method}

\subsection{PhysRepair: Generate--Diagnose--Refine Framework}
\label{sec:framework}

PhysRepair is an agentic video generation framework that turns video synthesis
from a one-pass procedure into an iterative
\emph{generate--diagnose--refine} process. Given a generation condition $c$,
a base video generator first produces an initial video
\begin{equation}
V^{(0)} = G_{0}(c).
\end{equation}
Rather than treating $V^{(0)}$ as the final output, PhysRepair regards it as
an intermediate generation whose realized dynamics can be explicitly inspected.
At each iteration $k$, a Physics Repair Agent diagnoses the current video
$V^{(k)}$, constructs a targeted physical correction, executes a localized
counterfactual repair, and verifies the resulting video. This feedback-driven
process allows physical guidance to be determined from the failure that
actually emerges in the generated video, rather than being fully specified
before synthesis.

\begin{figure}[t]
    \centering
    \includegraphics[width=\columnwidth]{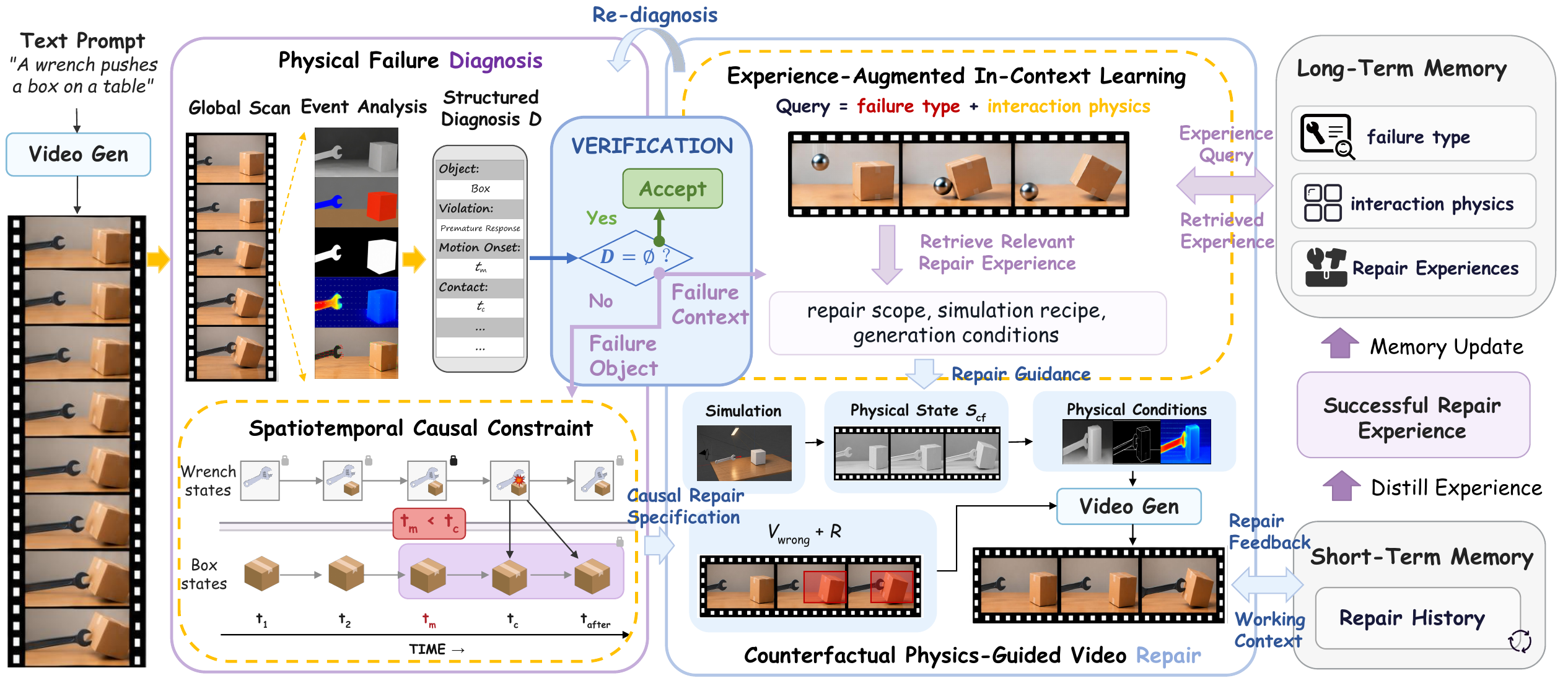}
    \caption{
    \textbf{Overview of PhysRepair.}
    Starting from an initially generated video, a Physics Repair Agent diagnoses the realized physical failure and performs localized counterfactual repair.
    Spatiotemporal causal constraints determine \emph{what may change and what must remain fixed}, while experience-augmented in-context learning retrieves relevant repair experience to adapt \emph{how to repair}.
    The resulting counterfactual physical state guides video regeneration, followed by physics-and-preservation verification and experience distillation for subsequent repair.
    }
\end{figure}

\paragraph{Evidence-grounded physical diagnosis.}

Given the current generation $V^{(k)}$, the agent first performs a coarse global scan using sparsely sampled frames to identify suspicious physical events and approximately localize where and when a potential violation occurs. It then increases the temporal resolution around the localized event window and selectively invokes perception tools for tracking \citep{karaev2025cotracker3}, segmentation \citep{ravi2024sam2}, optical flow \citep{lai2026cowtracker} and depth \citep{lin2025depth}. The resulting fine-grained evidence is used to determine the affected entities, failure type, and precise temporal relation, yielding a structured diagnosis $D^{(k)}$. This evidence-grounded diagnosis provides the basis
for subsequent repair while avoiding unnecessary modification of already
valid content.

\paragraph{Counterfactual physics-guided repair.}
Given a structured diagnosis $D^{(k)}$, PhysRepair must first determine
\emph{what to repair}. To this end, it constructs
a localized causal repair specification $R^{(k)}$, which identifies the
interaction to be corrected, the editable spatiotemporal region, and the
content that should remain unchanged. We detail this construction in
Sec.~\ref{sec:causal_constraint}.

Given $R^{(k)}$, the framework then determines \emph{how to repair} the
diagnosed interaction by adapting the repair strategy to the current failure.
This decision is guided by the Experience-Augmented In-Context Learning
mechanism, which retrieves relevant prior repair experience and provides it
in context to the Physics Repair Agent, as detailed in
Sec.~\ref{sec:experience_icl}. Based on the resulting repair strategy,
PhysRepair reconstructs only the local physical scene required to resolve the
diagnosed interaction. 
An executable physics simulator $\mathrm{Sim}(\cdot)$ \citep{coumans2021, blender2018}
then evolves this local scene to obtain a physically admissible
counterfactual state
\begin{equation}
    S_{\mathrm{cf}}^{(k)}
    =
    \mathrm{Sim}\!\left(R^{(k)}\right),
\end{equation}
which describes the corrected physical evolution rather than the final
photorealistic appearance.

PhysRepair transfers this physical correction back to video generation through two parallel branches. 
First, residual physically inconsistent dynamics in the source video are suppressed within the editable region, producing a decontaminated source $V_{\mathrm{clean}}^{(k)}$ while preserving valid appearance and unaffected content.
In parallel,
$S_{\mathrm{cf}}^{(k)}$ is converted into generation-compatible physical
conditions $C_{\mathrm{phys}}^{(k)}$, including trajectory, depth, motion
flow, and contact or mask cues. The candidate repair is
\begin{equation}
    V^{(k+1)}
    =
    G_{\theta}\!\left(
        V_{\mathrm{clean}}^{(k)},
        C_{\mathrm{phys}}^{(k)},
        R^{(k)}
    \right).
\end{equation}

\paragraph{Verification and iterative refinement.}
Verification is implemented as target-conditioned re-diagnosis: each repaired candidate is re-observed by the same evidence-grounded diagnostic process, and any residual or newly introduced violation directly initializes the next repair iteration. A successful repair is accepted; otherwise,
the verification outcome is returned to the Physics Repair Agent for
another targeted refinement.

This loop defines the execution backbone of PhysRepair. Two decisions remain
to instantiate targeted and adaptive repair:  \emph{what should be
modified} after a physical failure is diagnosed, and \emph{how the repair
process should adapt} across different failures and iterative attempts. We address these decisions with Spatiotemporal Causal Constraints and Experience-Augmented In-Context
Learning, respectively.

\subsection{Spatiotemporal Causal Constraints}
\label{sec:causal_constraint}

The framework above relies on a repair specification $R^{(k)}$ to localize
each intervention, but a structured diagnosis $D^{(k)}$ alone does not
determine which states should be modified and which should remain fixed.
We therefore derive the repair boundary from the causal structure of the
diagnosed interaction. Specifically, we organize the relevant object states into an object-centric spatiotemporal graph
\begin{equation}
    G^{(k)}=(\mathcal{V}^{(k)},
    \mathcal{E}_{\mathrm{temp}}^{(k)},
    \mathcal{E}_{\mathrm{int}}^{(k)}).
\end{equation}
Here, $\mathcal{E}_{\mathrm{temp}}$ connects successive states of the same
object, while $\mathcal{E}_{\mathrm{int}}$ represents cross-object physical
relations such as contact, collision, and support. The graph therefore
captures both object-wise temporal evolution and physical interactions
between objects.

Using $D^{(k)}$, we locate the first physically invalid state within the
diagnosed interaction and trace its causally affected descendants through
the graph. Only these affected states are marked as editable, while
unrelated object states, trajectories, and scene content are preserved.
The editable support is consequently localized jointly over object, space,
and time rather than by a spatial mask or temporal interval alone. This yields the causal repair specification
\begin{equation}
    R^{(k)}
    =
    \left(
        \mathcal{I}^{(k)},
        \mathcal{E}^{(k)},
        \mathcal{P}^{(k)}
    \right),
\end{equation}
where $\mathcal{I}^{(k)}$ denotes the intervention target,
$\mathcal{E}^{(k)}$ the editable object--space--time region, and
$\mathcal{P}^{(k)}$ the states and visual content to be preserved.
$R^{(k)}$ serves both as the specification of the local interaction for
counterfactual simulation and as the constraint for suppressing contaminated
source dynamics during video reconditioning.

\subsection{Experience-Augmented In-Context Learning}
\label{sec:experience_icl}

The causal repair specification determines where to intervene, but not how
the repair strategy should adapt across diverse physical failures. Different
failures may require different evidence, tool selections, simulation
configurations, generation conditions, and recovery strategies. We therefore
augment the Physics Repair Agent with in-context repair experience at both
the current-sample and cross-sample levels.

For the current video, a short-term working memory $\mathcal{M}_s$ maintains
the evolving repair context, including observed evidence, current hypotheses,
repair history, and verification feedback. A long-term experience memory
$\mathcal{M}_l$ stores successful repairs and reusable
repair skills accumulated across previous videos. Retrieved experience is
provided in context to the agent together with the current working state,
rather than being directly injected into the video generator.

At each decision step $t$, the agent retrieves relevant repair experience from
the long-term experience memory based on the current failure context $q_t$,
and selects the next action according to
\begin{equation}
    a_t
    =
    \pi\!\left(
        o_t,
        \mathcal{M}_s,
        \operatorname{Retrieve}(\mathcal{M}_l,q_t)
    \right),
\end{equation}
where $o_t$ denotes the current observation, $pi$ denotes the agent policy that selects repair actions based on the current observation, memory state, and retrieved experience and $a_t$ corresponds to
analysis, intervention, execution, verification, or backtracking.
Specifically, the retrieval query $q_t$ is constructed from the diagnosed
failure type and interaction physics that characterize the current repair
scenario. The similarity between $q_t$ and stored experiences is measured in
the embedding space, and the most relevant repair trajectories are retrieved
to provide in-context guidance. These retrieved experiences help the agent
select appropriate repair operations, recover from unsuccessful attempts, and
refine future decisions through accumulated repair history.

Verification feedback updates $\mathcal{M}_s$ for subsequent refinement, while successful repair experiences are selectively distilled into $\mathcal{M}_l$ as cross-sample experience. These include verified corrections of individual sub-errors during iterative repair, even when the overall failure remains unresolved, provided that their repair strategies differ from the retrieved experiences. The distilled experiences preserve the failure context, repair strategy, and verification outcome for future reuse.

\section{Experiments}
\label{sec:experiments}

We evaluate whether PhysRepair improves physical plausibility while preserving
content outside the diagnosed repair scope. We further study the contribution
of its causal constraint and experience-augmented repair strategy.

\subsection{Experimental Setup}
\label{sec:exp_setup}

\paragraph{Benchmark and metrics.}
We evaluate on \textbf{Physics-IQ Verified}
\citep{motamed2025physicsiq,radsch2026physicsiqverified}, comprising 198
take-1 instances from 66 real-world physical experiments observed from three
viewpoints. The benchmark spans solid mechanics, fluid dynamics, optics,
thermodynamics, and magnetism. We use the verified physical descriptions and
switch frames for generation, with model-specific prompt formatting following
the benchmark protocol when available. Reference continuations and evaluator
masks are never exposed to PhysRepair or its experience memory and are accessed
only during evaluation. We follow the official evaluation
protocol and report the aggregate \textbf{Physics-IQ Verified} score. The evaluator additionally measures
Spatial IoU, Spatiotemporal IoU, Weighted Spatial IoU, and RGB MSE, which
jointly characterize the spatial and temporal agreement between generated
videos and physically valid reference dynamics.

\setcounter{topnumber}{1}

\begin{table*}[t]
    \centering
    \caption{\textbf{Quantitative results on Physics-IQ Verified.} SP, ST, and WS denote Spatial IoU,
    Spatiotemporal IoU, and Weighted Spatial IoU, respectively. The best, second-best,
    and third-best results for each metric are marked in \textbf{bold},
    \underline{single underline}, and \uuline{double underline}, respectively.}
    \label{tab:physics_iq_verified_main}
    \resizebox{\textwidth}{!}{%
    \begin{tabular}{lccccc}
        \toprule
        \textbf{Method} & \textbf{SP}$\uparrow$ & \textbf{ST}$\uparrow$ & \textbf{WS}$\uparrow$ &
        \textbf{MSE score}$\uparrow$ & \textbf{Physics-IQ}$\uparrow$ \\
        \midrule
        \multicolumn{6}{c}{\textit{Video generation models}} \\
        \midrule
        MAGI-1 4.5B \citep{sandai2025magi1}
            & 14.3 & 13.3 & 7.0 & -- & 18.8 \\
        Cosmos-Predict2-2B \citep{nvidia2025cosmos}
            & -- & -- & -- & -- & 27.7 \\
        Wan2.2-TI2V-5B \citep{wan2025wan}
            & 35.9 & 26.4 & 23.0 & 25.5 & 27.7 \\
        CogVideoX-I2V-5B \citep{yang2024cogvideox}
            & 37.8 & \underline{35.5} & 21.8 & 32.0 & 31.8 \\
        Wan2.2-I2V-14B \citep{wan2025wan}
            & 51.1 & 20.5 & 28.5 & 28.9 & 32.2 \\
        HunyuanVideo-1.5-I2V \citep{wu2025hunyuanvideo15}
            & 47.1 & 26.9 & 29.7 & 30.0 & 33.4 \\

        \rowcolor{gray!10}
        MiniMax H3 FL2VA
            & \textbf{58.9} & 22.8 & \uuline{40.5} & \uuline{37.0} & \uuline{39.8} \\

        Cosmos3-Super-Image2Video \citep{nvidia2026cosmos3}
            & \uuline{57.4} & 32.5 & \underline{43.0} & \textbf{37.8} & \underline{42.7} \\
        \midrule
        \multicolumn{6}{c}{\textit{Physics-aware methods}} \\
        \midrule
        VLIPP \citep{yang2025vlipp}
            & 45.5 & 29.9 & 29.0 & 25.1 & 32.4 \\
        PhyCo \citep{narayanan2026phyco}
            & 46.2 & \textbf{43.8} & 30.5 & 26.8 & 36.8 \\
        \midrule

        \rowcolor{gray!10}
        \textbf{PhysRepair (Ours)}
            & \underline{58.2} & \uuline{34.7} & \textbf{44.2} & \underline{37.4} & \textbf{43.6} \\

        \bottomrule
    \end{tabular}%
    }
\end{table*}

\begin{figure}[t]
    \centering
    \includegraphics[width=\linewidth]{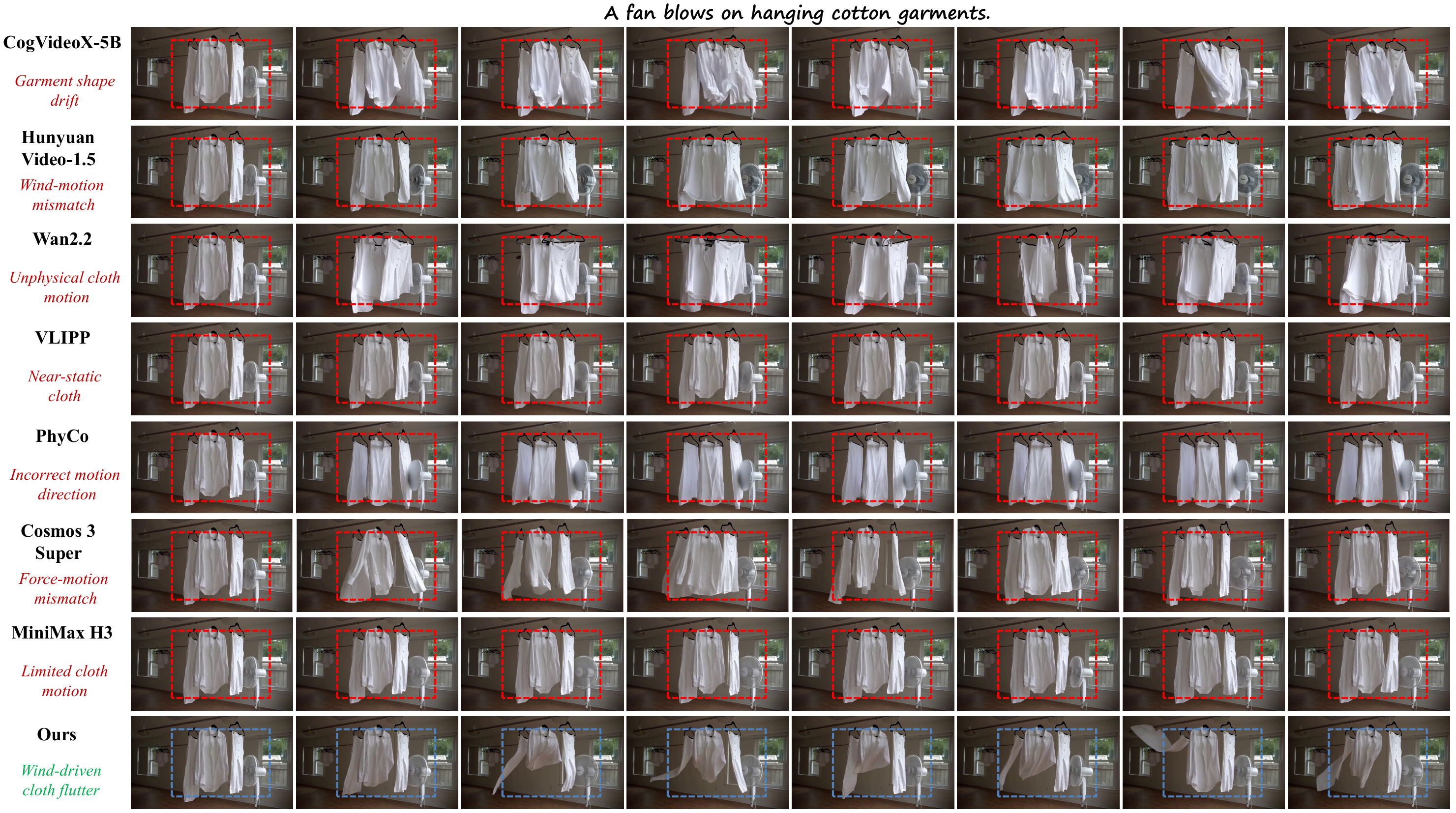}\\[-2pt]
    \includegraphics[width=\linewidth]{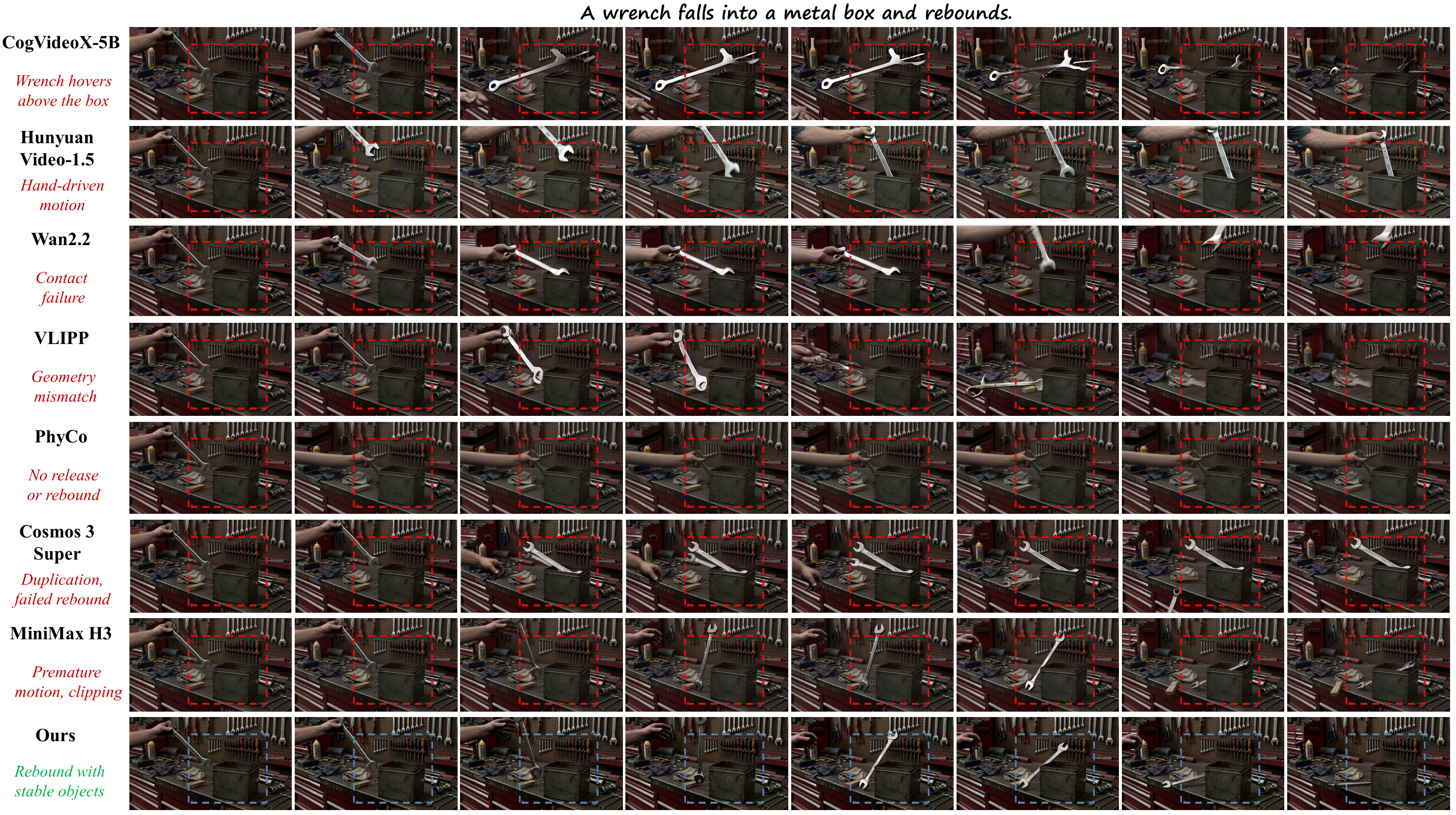}
    \caption{\textbf{Qualitative comparison on multi-object spatiotemporal interactions.}
    Top: fan–garment interaction; bottom: wrench–box interaction.
    Baselines exhibit incorrect contact timing, inconsistent motion, or object artifacts, whereas PhysRepair preserves object identity and coherent contact--response dynamics.}
    \label{fig:qualitative_comparison}
\end{figure}

\paragraph{Baselines.}
We compare against six general-purpose video generators: \textbf{Wan2.2}
\citep{wan2025wan}, \textbf{CogVideoX-I2V-5B}
\citep{yang2024cogvideox}, \textbf{Cosmos-Predict2-2B}
\citep{nvidia2025cosmos}, \textbf{Cosmos3-Super-Image2Video}
\citep{nvidia2026cosmos3}, \textbf{MAGI-1 4.5B}
\citep{sandai2025magi1}, and \textbf{HunyuanVideo-1.5-I2V}
\citep{wu2025hunyuanvideo15}, as well as two physics-aware approaches:
\textbf{VLIPP} \citep{yang2025vlipp} and \textbf{PhyCo}
\citep{narayanan2026phyco}. Every method receives the same switch
frame and prompt, and failed samples are retained in the evaluation.

\paragraph{Implementation details.}
PhysRepair generates and repairs videos using the frozen
MiniMax H3 Ref2VA backbone. Specifically, the initial video is generated using MiniMax H3 FL2VA. GPT-6-Astra performs structured diagnosis, causal
planning, verification, and experience distillation. Perception and
registration combine SAM~2 \citep{ravi2024sam2} with tracking \citep{karaev2025cotracker3}, optical flow \citep{lai2026cowtracker} and depth \citep{lin2025depth} tools, while the diagnosed local interaction is reconstructed and simulated
in simulator \citep{coumans2021, blender2018}. The resulting counterfactual state is converted into
generation-compatible structural and motion conditions and injected through
the H3 control adapter. The generator uses 20 denoising steps, and the agent
performs at most 20 repair rounds, terminating once the physics criteria are satisfied.

\subsection{Main Results}
\label{sec:main_results}

\paragraph{Quantitative comparison on Physics-IQ.}
Table~\ref{tab:physics_iq_verified_main} compares PhysRepair with
general-purpose video generators and physics-aware methods. PhysRepair
achieves the highest overall Physics-IQ score of 43.6 among the reported
methods, surpassing the strongest video-generation baseline,
Cosmos3-Super-Image2Video, by 0.9 points. Compared with MiniMax H3 FL2VA,
PhysRepair improves Physics-IQ from 39.8 to 43.6, demonstrating the benefit
of targeted physical repair over the initial generation. The improvement is concentrated in the spatiotemporal metrics. In particular,
PhysRepair increases ST from 22.8 to 34.7 (+11.9) and WS from 40.5 to 44.2
(+3.7) over MiniMax H3 FL2VA. This pattern is
consistent with the design of PhysRepair: rather than globally modifying
already-valid visual content, it localizes causally affected states and
corrects the corresponding interaction dynamics. The larger gains in ST and
WS therefore indicate improved spatiotemporal agreement with physically valid
dynamics, while the stable SP and MSE scores suggest that these corrections
are achieved without substantially perturbing the original spatial appearance.

\begin{figure}[t]
    \centering
    \includegraphics[width=\columnwidth]{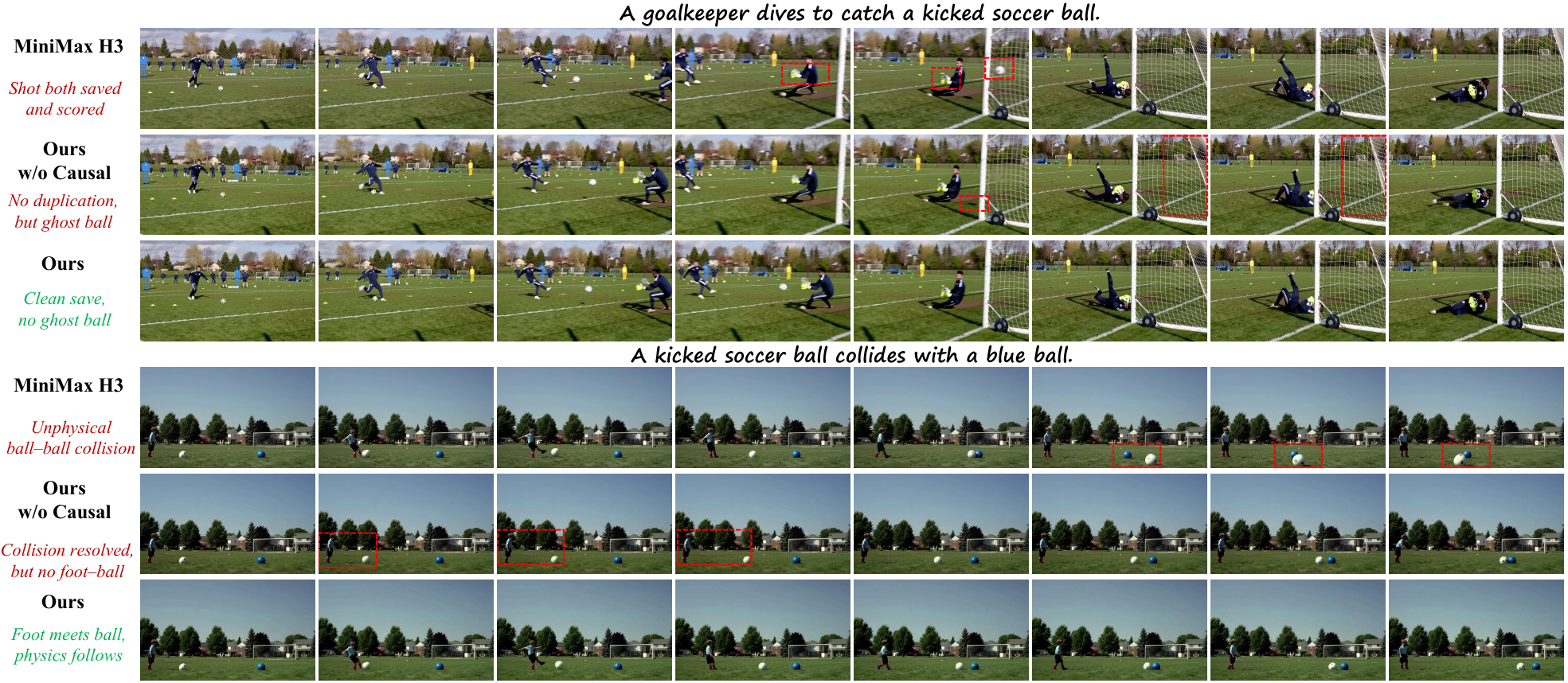}\\[-2pt]
    \includegraphics[width=\linewidth]{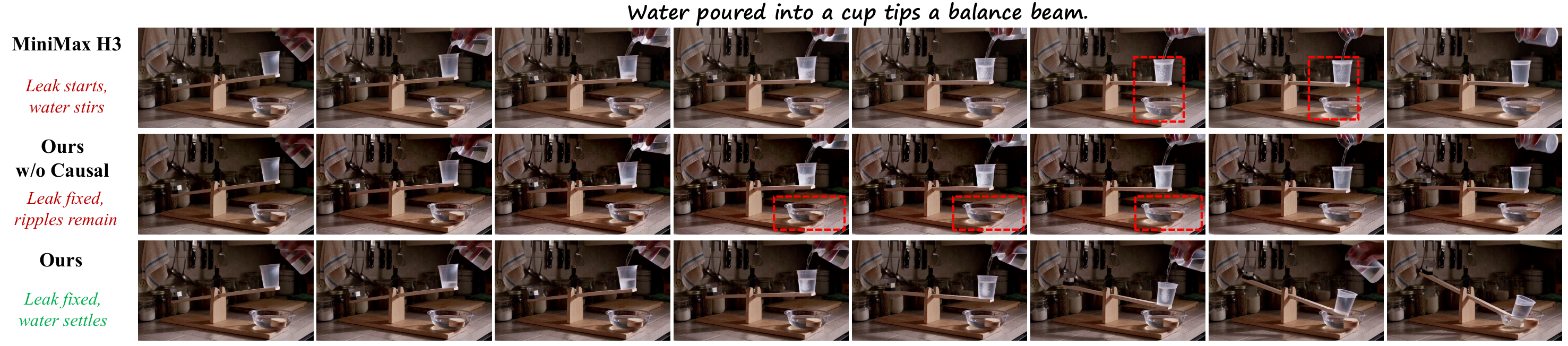}
    \caption{\textbf{Qualitative ablation of Spatiotemporal Causal Constraints.}
    Without causal constraints, repair may correct the immediate target while leaving causally affected states unresolved or disrupting valid interactions. Full PhysRepair jointly constrains editable and preserved states across the interaction, enabling complete and localized physical repair.}
    \label{fig:causal_ablation}
\end{figure}

\begin{figure}[t]
    \centering

    \begin{subfigure}[t]{0.50\columnwidth}
        \centering
        \includegraphics[width=\linewidth]{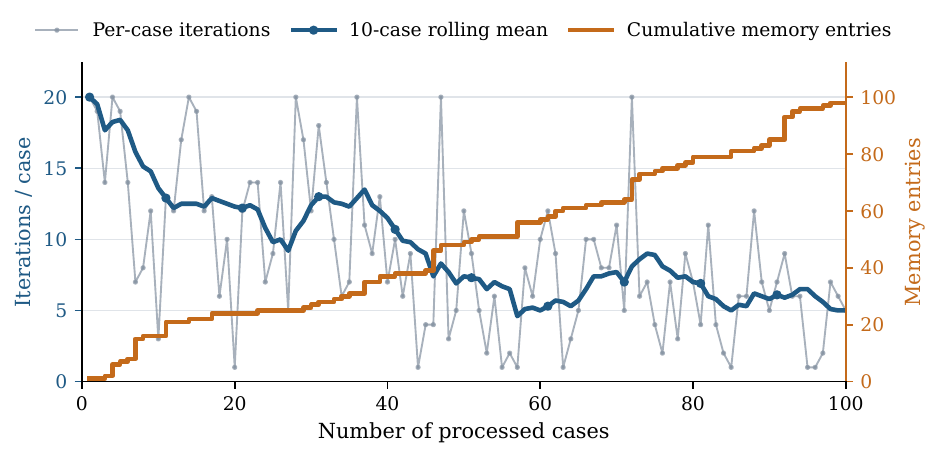}
        \caption{
        \textbf{Long-term memory growth and repair efficiency over the synthetic-case
        sequence.}
        The gray curve shows per-case repair iterations, the blue curve
        shows their ten-case rolling mean, and the orange staircase shows cumulative
        long-term memory entries.
        }
        \label{fig:icl_memory_efficiency}
    \end{subfigure}
    \hfill
    \begin{subfigure}[t]{0.46\columnwidth}
        \centering
        \includegraphics[width=\linewidth]
        {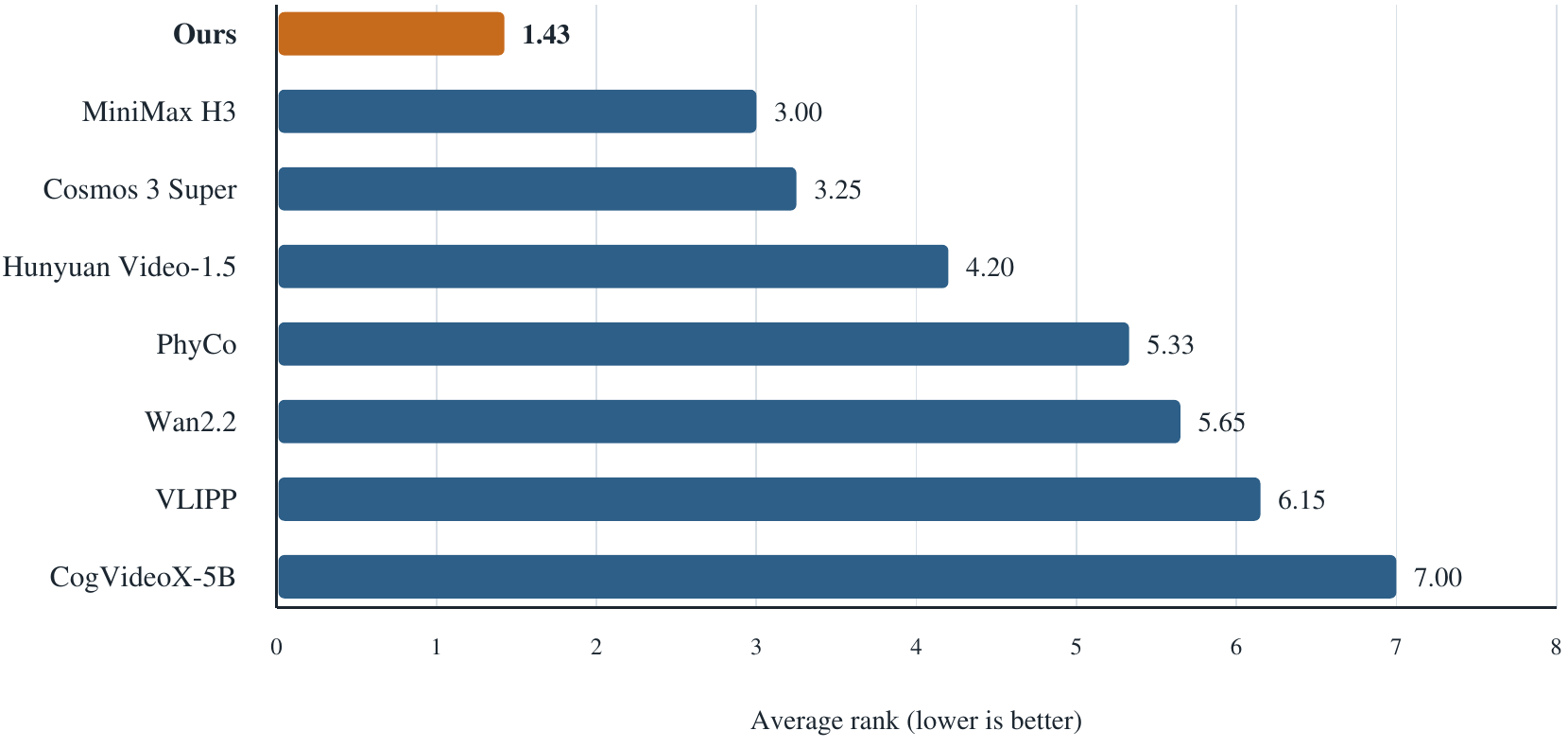}
        \caption{
        \textbf{User study on perceived physical consistency.}
        PhysRepair achieves the best average rank among eight compared methods,
        indicating that users consistently perceive its generated videos as more
        physically plausible.
        }
        \label{fig:user_study}
    \end{subfigure}

    \caption{\textbf{Analysis of PhysRepair.}
    Memory efficiency and user-perceived physical consistency of PhysRepair.}
    \label{fig:analysis}
\end{figure}

\paragraph{Qualitative comparison on multi-object spatiotemporal interactions.}
Figure~\ref{fig:qualitative_comparison} further compares different methods on two contact-driven interactions that require maintaining physical consistency across multiple interacting objects and over time. In the fan--garment example, the generated sequence must preserve the identities of the fan and the hanging garments, establish the direction of the fan-induced airflow, and produce garment motion consistent with the resulting aerodynamic force. Existing methods frequently break this interaction through weak or missing force responses despite visible fan motion (VLIPP, MiniMax H3), physically implausible garment motion (Wan2.2), or garment responses whose motion direction is inconsistent with the fan-induced airflow (Cosmos3-Super). In contrast, PhysRepair preserves the identities of both the fan and the garments and restores the correct force--response dynamics: the fan first generates directed airflow, after which the garments deform and sway in a physically plausible manner consistent with the applied force. This result illustrates how our Spatiotemporal Causal Constraints support targeted correction of causally affected states while preserving valid object states, enabling coherent force-driven dynamics without unnecessary changes to unaffected content.

The wrench--box example requires a longer interaction sequence: the wrench falls toward the box, makes contact, and subsequently rebounds. Competing methods often realize only part of this process, e.g., failing to establish effective contact (Wan2.2) or introducing substantial geometry and trajectory inconsistencies (VLIPP). More specifically, MiniMax H3 exhibits a spatiotemporal causal misalignment in which the target box begins moving before the actual collision, together with visible wrench--box interpenetration, whereas Cosmos 3 Super leaves the wrench stalled at the box instead of completing the expected rebound. In contrast, PhysRepair preserves the wrench and box identities while restoring a coherent fall--contact--rebound sequence. This result illustrates how causally constrained, simulation-guided repair can address both incorrect contact timing and subsequent motion inconsistencies while preserving unaffected content. Together, these cases highlight the benefit of identifying causally affected states and guiding their localized regeneration with physically plausible counterfactual dynamics, rather than correcting individual frames or objects in isolation.

\paragraph{User Study.} We conduct a user study in which participants rank videos generated by the compared methods according to their perceived physical consistency. As shown in Fig.~\ref{fig:user_study}, PhysRepair achieves the best average rank among all compared methods, indicating that its improvements are also reflected in human judgments of physical plausibility. These results complement the quantitative evaluation and support the effectiveness of our \textit{generate--diagnose--repair} paradigm in producing perceptually more convincing physical dynamics.

\subsection{Ablation and Analysis}
\label{sec:ablation}

\paragraph{Spatiotemporal Causal Constraints.}
We qualitatively examine how Spatiotemporal Causal Constraints affect localized physical repair. As shown in Fig.~\ref{fig:causal_ablation}, without the causal constraint, the agent can often correct the most salient error on the directly involved object, but may leave causally affected states unresolved or alter valid interactions that should remain fixed. In the goalkeeper example, MiniMax H3 produces a contradictory outcome in which the shot is both saved and scored. Without the causal constraint, the duplicated ball is removed, but a residual ball artifact in the goal and continued net motion still imply a scoring event. The full model additionally corrects these downstream states, yielding a clean save. In the ball-collision example, the source video exhibits an implausible collision between the soccer ball and the blue ball. Removing the causal constraint resolves the collision but disrupts the preceding foot--ball contact, causing the ball to move without a valid kick. In contrast, the full model preserves the valid kick while correcting the subsequent collision dynamics.  In the water-leakage example, the source video exhibits spurious water leakage from the upper cup into the basin below, which consequently induces fluid motion in the basin. Without the causal constraint, the leakage itself is corrected, but the resulting motion of the water in the basin remains, leaving a downstream effect without its corresponding cause. In contrast, the full model propagates the correction to these causally dependent states and removes the residual fluid response, restoring a physically consistent interaction. Together, these cases show that causal constraints are important for both propagating corrections to downstream affected states and preserving valid states outside the repair scope, enabling complete yet localized physical repair.

\paragraph{Experience-Augmented In-Context Learning.}
For this analysis, we select 100 cases from the videos initially generated by MiniMax H3 FL2VA, each diagnosed as containing a physical inconsistency. These cases are processed sequentially to examine how repair efficiency evolves as long-term experience memory accumulates. We examine how accumulated cross-sample experience affects iterative repair efficiency. As shown in Fig.~\ref{fig:icl_memory_efficiency}, the moving average of repair rounds decreases as long-term experience memory grows, despite variation across individual cases. Difficult cases can still exhaust the 20-round budget, but such cases become less frequent as experience accumulates. The results suggest that experience reuse can reduce repeated exploration and improve repair efficiency without updating model weights. This trend is consistent with the intended role of our Experience-Augmented In-Context Learning: combining current repair feedback with relevant prior experience to guide intervention selection and recovery from unsuccessful attempts.

\section{Limitations and Broader Impact}
\label{sec:limitations}

\paragraph{Limitations.}
PhysRepair currently relies on a capable multimodal language model as the
repair agent for physical diagnosis, tool selection, causal reasoning,
verification, and experience reuse. Consequently, its performance is partly
bounded by the agent model's physical reasoning and tool-use capabilities,
and weaker agent backbones may lead to less reliable repair decisions.
This dependence also introduces additional inference cost compared with
one-pass video generation, particularly when multiple repair rounds are
required.

\paragraph{Broader Impact.}
PhysRepair provides a general mechanism for improving the physical
plausibility of generated videos without retraining the underlying video
generator, which may benefit applications that require more reliable physical
dynamics, such as simulation, embodied AI, and synthetic data generation.
At the same time, increasingly realistic physical video generation and repair
may make synthetic content more difficult to distinguish from real recordings,
potentially increasing risks of deceptive or misleading media. We therefore
view provenance mechanisms and responsible deployment practices as important
complements to further improvements in physically grounded video generation.

\section{Conclusion}
\label{sec:conclusion}

We presented PhysRepair, an agentic framework for improving the physical plausibility of generated videos through an iterative generate–diagnose–repair process. Rather than imposing physical guidance entirely before synthesis, PhysRepair diagnoses the realized failure and uses Spatiotemporal Causal Constraints to localize the causally affected states. Under this constraint, the Physics Repair Agent derives a physically admissible counterfactual evolution for the affected interaction, which guides localized repair while preserving unaffected content. Experience-augmented In-context Learning further adapts repair decisions using current-sample feedback and relevant experience from previous repairs. Experiments on Physics-IQ Verified across five physics categories demonstrate consistent improvements in physical plausibility, suggesting post-hoc adaptive repair as a complementary direction for physically grounded video generation.

\bibliographystyle{iclr2027_conference}
\bibliography{references}
\end{document}